%% file: 0_main.tex
\documentclass[10pt,twocolumn,letterpaper]{article}

\usepackage{cvpr}
\usepackage{times}
\usepackage{epsfig}
\usepackage{graphicx}
\usepackage{amsmath}
\usepackage{amssymb}
\usepackage{mathrsfs}
\usepackage{calligra}

\usepackage{tabulary,multirow,overpic,xcolor}
\usepackage[caption=false]{subfig}

\usepackage[english]{babel}

\newcolumntype{x}[1]{>{\centering\arraybackslash}p{#1pt}}

\newcommand{\app}{\raise.17ex\hbox{$\scriptstyle\sim$}}

\newlength\savewidth\newcommand\shline{\noalign{\global\savewidth\arrayrulewidth
  \global\arrayrulewidth 1pt}\hline\noalign{\global\arrayrulewidth\savewidth}}
\newcommand{\tablestyle}[2]{\setlength{\tabcolsep}{#1}\renewcommand{\arraystretch}{#2}\centering\footnotesize}
\newcommand{\myparagraph}[1]{{\vspace{0.5em} \noindent \bf #1}}

\makeatletter\renewcommand\paragraph{\@startsection{paragraph}{4}{\z@}
  {.5em \@plus1ex \@minus.2ex}{-.5em}{\normalfont\normalsize\bfseries}}\makeatother

\definecolor{citecolor}{RGB}{34,139,34}
\usepackage[pagebackref=true,breaklinks=true,letterpaper=true,colorlinks, citecolor=citecolor,bookmarks=false]{hyperref}

\cvprfinalcopy %

\def\cvprPaperID{6980} %

\ifcvprfinal\fi

\begin{document}

\title{Something-Else: Compositional Action Recognition with \\ Spatial-Temporal Interaction Networks}

\author{ Joanna Materzynska\\
University of Oxford, TwentyBN
\and
Tete Xiao\\
UC Berkeley\\
\and
Roei Herzig\\
Tel Aviv University\\
\and
Huijuan Xu\footnotemark[1]\\
UC Berkeley\\
\and
Xiaolong Wang\footnotemark[1]\\
UC Berkeley\\
\and
Trevor Darrell\thanks{Equal advising}\\
UC Berkeley\\
}

\maketitle
\thispagestyle{empty}

\begin{abstract}
\vspace{-0.05in}
Human action is naturally compositional: humans can easily recognize and perform actions with objects that are different from those used in training demonstrations. In this paper, we study the compositionality of action by looking into the dynamics of subject-object interactions. We propose a novel model which can explicitly reason about the geometric relations between  constituent objects and an agent performing an action. To train our model, we collect dense object box annotations on the Something-Something dataset. We propose a novel compositional action recognition task where the training combinations of verbs and nouns do not overlap with the test set.
The novel aspects of our model are applicable to activities with prominent object interaction dynamics and to objects which can be tracked using state-of-the-art approaches; for activities without clearly defined spatial object-agent interactions, we rely on baseline scene-level spatio-temporal representations. We show the effectiveness of our approach not only on the proposed compositional action recognition task, but also in a few-shot compositional setting which requires the model to generalize across both object appearance and action category.\footnote{Project page: https://joaanna.github.io/something\_else/.}
\vspace{-0.1in}
\end{abstract}

\input{1_introduction.tex}

\input{2_related.tex}

\input{3_approach.tex}

\input{4_experiments.tex}

\input{5_conclusion.tex}

{\noindent {\bf Acknowledgement}: Prof. Darrell's group was supported in part by DoD, NSF, BAIR, and BDD. We would like to thank Fisher Yu and Haofeng Chen for helping set up the  annotation pipeline, and Anna Rohrbach and Ronghang Hu for helpful discussions.}

{\small
\bibliographystyle{ieee_fullname}
\bibliography{egbib}
}

\end{document}

%% file: 1_introduction.tex
\vspace{-0.01in}
\section{Introduction}
\label{sec:intro}
\vspace{-0.03in}
\begin{figure}[t]
{
\centering

\includegraphics[clip,width=0.45\textwidth]{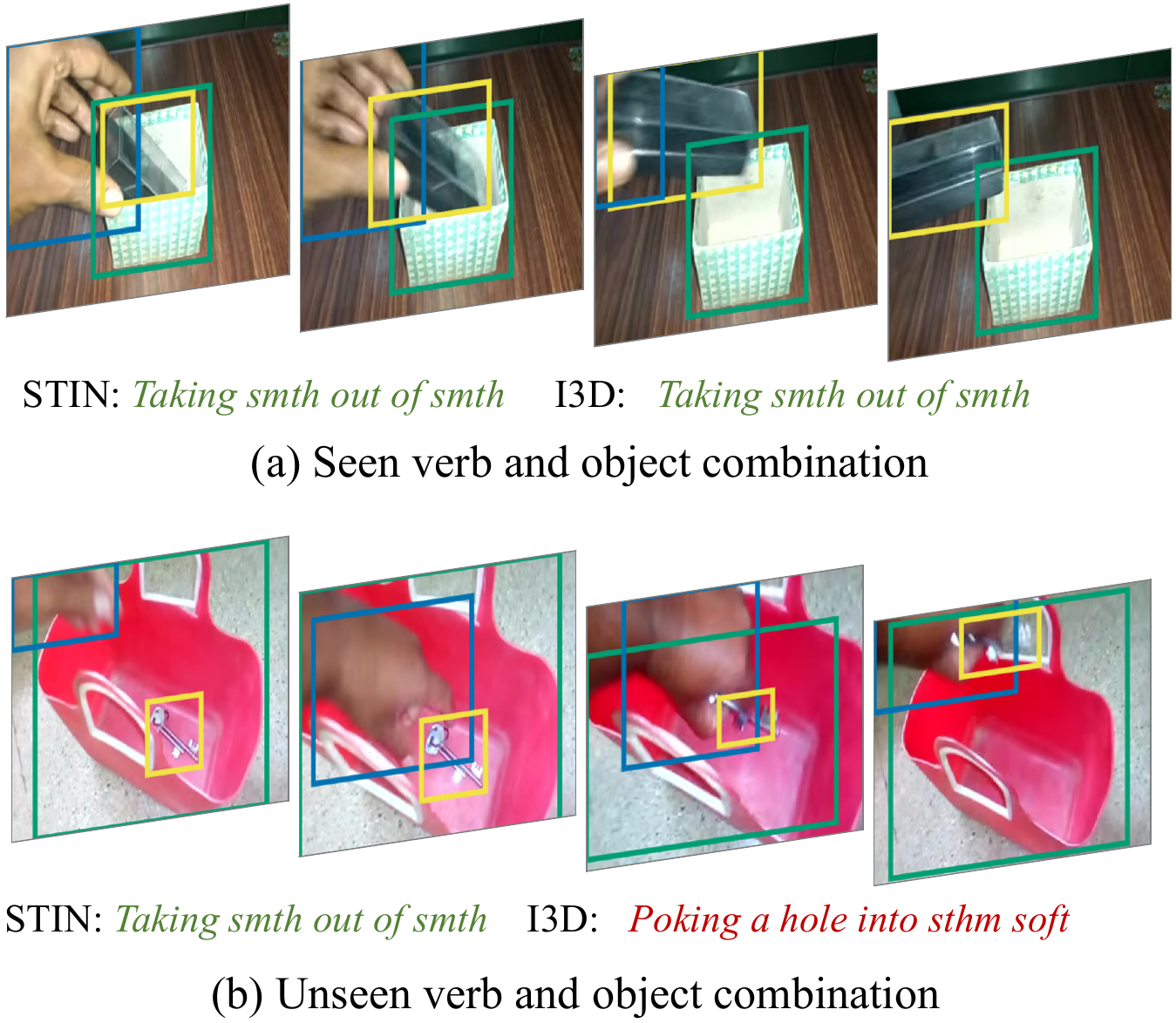}%
 \vspace{-0.05in}
 \caption{Two example videos of an action class ``taking something out of something'': the activity defines the relative change in object and agent (hand) positions over time. Most current methods (I3D-based) over-rely on object appearance. While it works well on seen verb and object combination in (a), it cannot generalize to unseen combinations in (b). Our Spatial-Temporal Interaction Networks (STIN) is designed for generalizing action recognition regardless of the object appearance in the training set. (Correct predictions are in green, incorrect in red.)
 } 
  \vspace{-0.05in}
\label{fig:teaser}
}
\end{figure}
\vspace{-0.01in}
Let's look at the simple action of ``taking something out of something'' in Figure~\ref{fig:teaser}. Even though these two videos show human hands interacting with different objects, we recognize that they are the same action based on changes in the relative positions of the objects and hands involved in the activity. Further, we can easily recognize the action even when it is presented with previously unseen objects and tools. We ask, do current machine learning algorithms have the capability to generalize across different combinations of verbs and nouns? \par
\vspace{-0.05in}
We investigate actions represented by the changes in geometric arrangements between subjects (agents) and objects. We propose a compositional action recognition setting in which we decompose each action into a combination of a verb, a subject, and one or more objects. Instead of the traditional setting where training and testing splits include the same combinations of verbs and nouns, we train and test our model on the same set of verbs (actions) but combine them with different object categories, so that tested verb and object combinations have never been seen during training time (Figure~\ref{fig:teaser} (b)). 
\par
\vspace{-0.03in}
This problem turns out to be very challenging for heretofore state-of-the-art action recognition models. Computer vision researchers have developed deep networks with temporal connections for action recognition by using Recurrent Neural Networks with 2D Convolutions~\cite{Yue-HeiNg2015,Donahue2015} and 3D ConvNets~\cite{Carreira2017,Xie17,Tran2015,Tran18}. However, both types of models have difficulty in this setting; our results suggest that they cannot fully capture the compositionality of action and objects.  These approaches focus on extracting features for the whole scene and do not explicitly recognize objects as individual entities; scene-level convolutional operators may rely more on spatial appearance rather than temporal transformations or geometric relations, since the former alone are often highly predictive of the action class~\cite{santoro2017simple,battaglia2018relational}.  
\par
\vspace{-0.03in}
Recently, researchers have investigated building spatial-temporal graph representations of videos~\cite{xiaolongwang2017nonlocal,wang2018videos,chen2019graph,Herzig_2019_ICCV} leveraging recently proposed graph neural networks~\cite{kipf2017semi}. These methods take dense object proposals as graph nodes and learn the relations between them. While this certainly opens a door for bringing relational reasoning in video understanding, the improvement over the 3D ConvNet baselines is not very significant. Generally, these methods have employed non-specific object graphs based on a large set of object proposals in each frame, rather than sparse semantically grounded graphs which model the specific interaction of an agent and constituent objects in an action.\par
\vspace{-0.03in}
In this paper, we propose a model based on a sparse and semantically-rich object graph learned for each action. We train our model with accurately localized object boxes in the demonstrated action. 
Our model learns explicit relations between subjects and objects; these turn out to be the key for successful compositional action recognition. We leverage state-of-the-art object detectors to accurately locate the subject (agent) and constituent objects in the videos, perform multi-object tracking on them and form multiple tracklets for boxes belonging to the same instance.  As shown in Figure~\ref{fig:teaser}, we localize the hand, and the objects manipulated by the hand. We track the objects over time and the objects belonged to the same instance are illustrated by the boxes with the same color.\par
\vspace{-0.03in}
Our Spatial-Temporal Interaction Network (STIN) reasons on candidate sparse graphs found from these detection and tracking results.
Our model takes the locations and shapes of objects and subject in each frame as inputs. It first performs spatial interaction reasoning on them by propagating the information among the subjects and objects, then
we perform temporal interaction reasoning over the boxes along the same tracklet, which encodes the transformation of objects and the relation between subjects and objects in time. Finally, we compose the trajectories for the agent and the objects together to understand the action. Our model is designed for activities which have prominent interaction dynamics between a subject or agent (\eg, hand) and constituent objects; for activities where no such dynamics are clearly discernible with current detectors (\eg, pouring water, crushing paper), our model falls back to leverage baseline spatio-temporal scene representations.

We introduce the Something-Else task, which extends the Something-Something dataset~\cite{goyal2017something} with new annotations and a new compositional split. In our compositional split, methods are required to recognize an action when performed with unseen objects, i.e., objects which do not appear together with this action at training time. Thus methods are trained on ``Something'', but are tested on their ability to generalize to ``Something-Else''. Each action category in this dataset is described as a phrase composed with the same verb and different nouns. We reorganize the dataset for compositional action recognition and model the dynamics of inter-object geometric configurations across time per action. We investigate compositional action recognition tasks in both a standard setting (where training and testing are with the same categories) and a few-shot setting (where novel categories are introduced with only a few examples). To support these two tasks, we collect and will release annotations on object bounding boxes for each video frame. Surprisingly, we observe even with only low dimensional coordinate inputs, our model can show comparable results and improves the appearance-based models in few-shot setting by a significant margin.

Our contributions include: (i) A Spatial-Temporal Interaction Network which explicitly models the changes of geometric configurations between agents and objects; (ii) Two new compositional tasks for testing model generalizability and dense object bounding box annotations in videos; (iii) Substantial performance gain over appearance-based model on compositional action recognition.

%% file: 2_related.tex
\vspace{-0.05in}
\section{Related Work}
\label{sec:related}
\vspace{-0.05in}

Action recognition is of central importance in computer vision. Over the past few years, researchers have been collecting larger-scale datasets including Jester~\cite{Materzynska_2019_ICCV},  UCF101~\cite{soomro2012ucf101}, Charades~\cite{Charades}, Sports1M~\cite{karpathy2014large} and Kinetics~\cite{Kinetics}. Boosted by the scale of data, modern deep learning approaches, including two-stream ConvNets~\cite{Simonyan2014,WangXWQLTV16}, Recurrent Neural Networks~\cite{Yue-HeiNg2015,Donahue2015LRCN,pan2016hierarchical,Bian2017} and 3D ConvNets~\cite{3DCNN,Carreira2017,Feichtenhofer2016,Xie2017,Tran15,Tran18,feichtenhofer2018slowfast}, have shown encouraging results on these datasets. However, a recent study in~\cite{zhou2018temporal} indicates that most of the current models trained with the above-mentioned datasets are not focusing on temporal reasoning but the appearance of the frames: Reversing the order of the video frames at test time will lead to almost the same classification result. In light of this problem, the Something-Something dataset~\cite{goyal2017something} is introduced to recognize action independent of the object appearance. To push this direction forward, we propose the compositional action recognition task for this dataset and provide object bounding box annotations.

The idea of compositionality in computer vision originates from Hoffman's research on Parts of Recognition~\cite{hoffman1984parts}. Following this work, models with pictorial structures have been widely studied in traditional computer vision~\cite{felzenszwalb2009object,zhu2007stochastic,ikizler2008searching}. For example, Felzenszwalb \etal~\cite{felzenszwalb2009object} proposes a deformable part-based model that organizes a set of part classifiers in a deformable manner for object detection. The idea of composing visual primitives and concepts has also been brought back in the deep learning community recently~\cite{abstractionTulsiani17,misra2017red,kato2018compositional,andreas2016neural,johnson2017inferring,hu2017learning}. For example, Misra \etal~\cite{misra2017red} propose a method to compose classifiers of known visual concepts and apply this model to recognize objects with unseen combinations of concepts. Motivated by this work, we propose to explicitly compose the subjects and objects in a video and reason about the relationships between them to recognize the action with unseen combinations of verbs and nouns.

The study of visual relationships has a long history in computer vision~\cite{Gupta09PAMI,Yao10CVPR,russell2006using} and early work investigated combining object and motion features for action recognition \cite{saenko2012mid,packer2012combined}.
Recent works have shown relational reasoning with deep networks on images~\cite{gkioxari2017interactnet,HuCVPR18,Santoro2017,johnson2015image}. For example, Gkioxari \etal~\cite{gkioxari2017interactnet} proposes to accurately detect the relations between the objects together with state-of-the-art object detectors. The idea of relational reasoning has also been extended in video understanding~\cite{xiaolongwang2017nonlocal,wang2018videos,wu2019long,Herzig_2019_ICCV,sun2018actor,girdhar2019video,Battaglia2016,watters2017visual,jain2016structural}. For instance, Wang \etal~\cite{wang2018videos} apply a space-time region graph to improve action classification in cluttered scenes. Instead of only relying on dense ``objectness'' region proposals, Wu \etal~\cite{wu2019long} further extend this graph model with accurate human detection and reasoning over a longer time range. Motivated by these works, we build our spatial-temporal interaction network to reason about the relations between subjects and objects based on accurate detection and tracking results. Our work is also related to the Visual Interaction Network~\cite{watters2017visual}, which models the physical interactions between objects in a simulated environment.

To further illustrate the generalizability of our approach, we also apply our model in a few-shot setting. Few-shot image recognition has become a popular research topic in recent years~\cite{finn2017model,snell2017prototypical,vinyals2016matching,Chen2019Closer,garcia2017few,ravi2016optimization}. 
Chen \etal~\cite{Chen2019Closer} has re-examined recent approaches in few-shot learning and found a simple baseline model which is very competitive compared to meta-learning approaches~\cite{finn2017model,lee2019meta,santoro2016meta}. 
Researchers have also investigated few-shot learning in videos~\cite{guo2018neural,cao2019few}. %
Guo \etal~\cite{guo2018neural} propose to perform KNN on object graph representations for few-shot 3D action recognition. We adopt our spatial-temporal interaction network for few-shot video classification, by using the same learning scheme as the simple baseline mentioned in~\cite{Chen2019Closer}.

%% file: 3_approach.tex
\vspace{-0.05in}
\section{Spatial-Temporal Interaction Networks}
\label{sec:stin}
\vspace{-0.05in}
\begin{figure}[t]
{
\centering
\includegraphics[clip,width=0.4\textwidth]{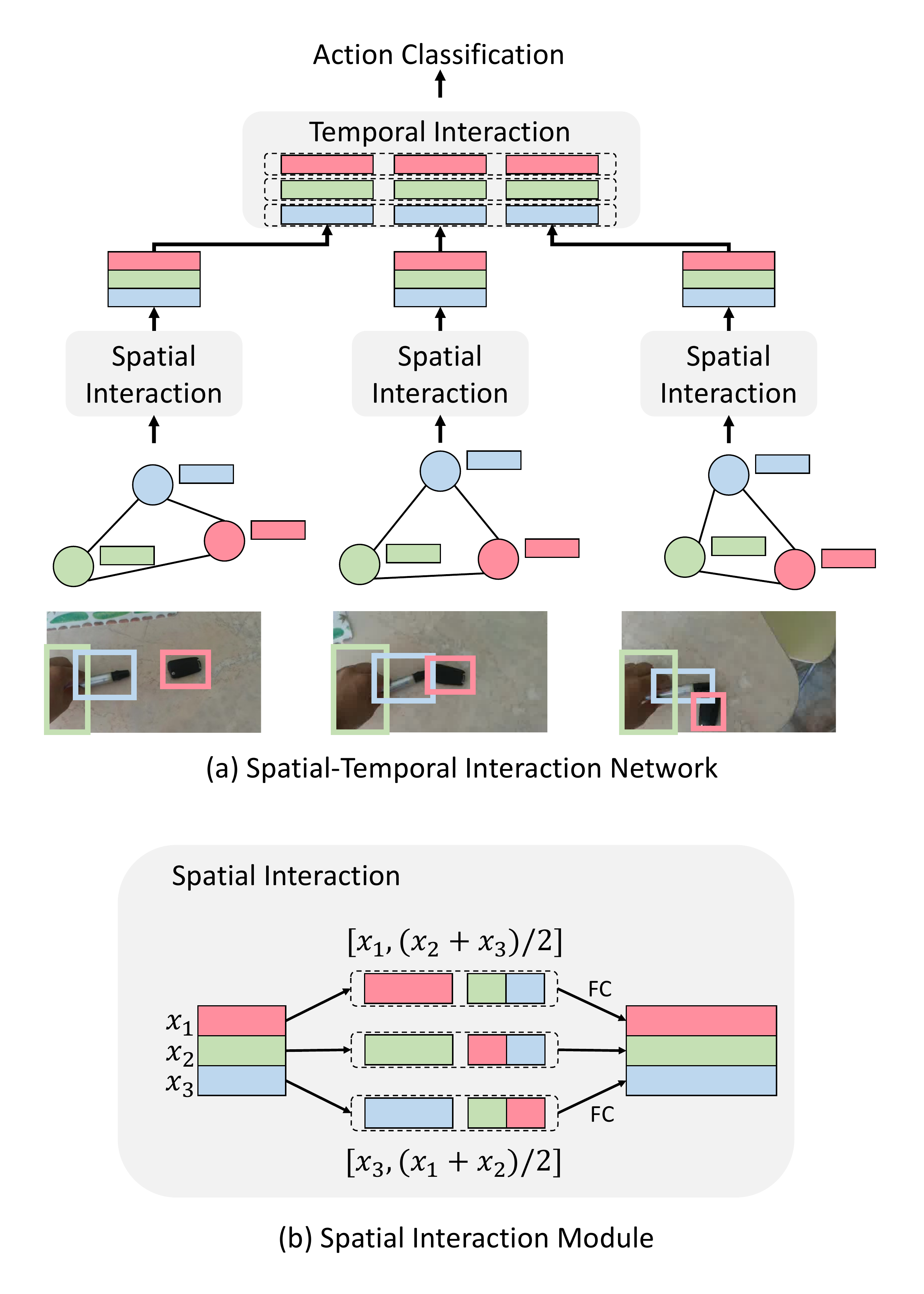}%
\vspace{-0.05in}
 \caption{(a) The Spatial-Temporal Interaction Network (STIN): our model operates on object-centric features and performs spatial interaction reasoning for individual frames and temporal interaction reasoning to obtain a  classification decision. (Different colors represent  different objects in this figure.) %
 (b) The Spatial Interaction Module: Given a set of object features in one frame, we aggregate them together with the information about their relative position by applying Eq.~\ref{eq:stm} to update each object feature. %
 }
 \vspace{-0.05in}
\label{fig:method}
}
\end{figure}
\begin{figure*}[t]
{
\centering
\includegraphics[width=0.95\linewidth]{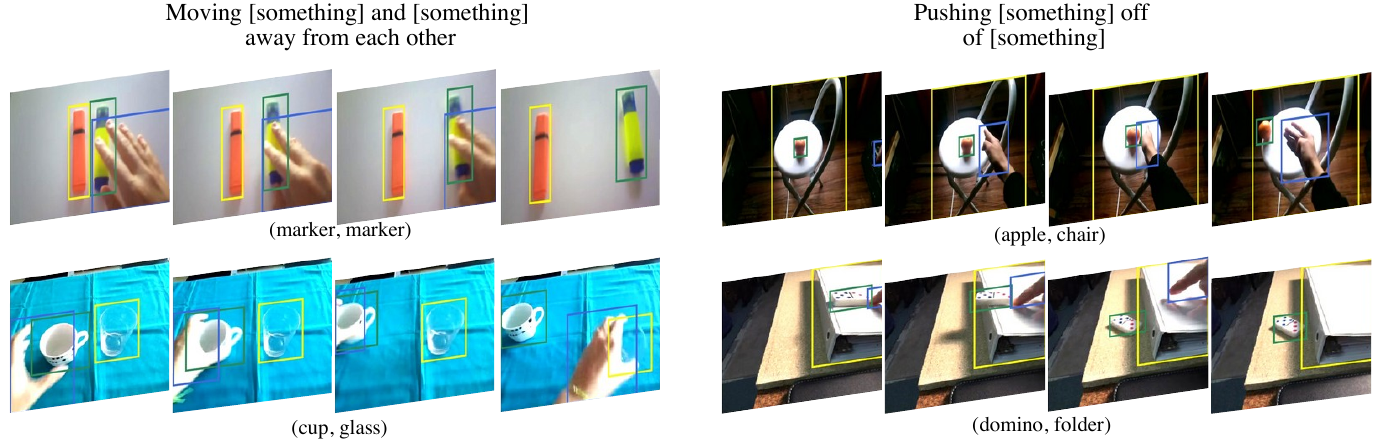}%
 \caption{Annotated examples of the Something-Something V2 dataset. Understanding the action from the visual appearance of the entire scene is challenging because we can perform the same action using arbitrary objects, however, observing the relative change of the location and positioning of the object and hands in the scene concisely captures the interaction.
 }
 \vspace{-0.1in}
\label{fig:dataset}
}
\end{figure*}

We present Spatial-Temporal Interaction Networks (STIN) for compositional action recognition. Our model utilizes a generic detector and tracker to build object-graph representations that explicitly include hand and constituent object nodes. We perform spatial-temporal reasoning among these bounding boxes to understand how the relations between subjects and objects change over time for a given action (Figure~\ref{fig:method}). By explicitly modeling the transformation of object geometric relations in a video, our model can effectively generalize to videos with unseen combinations of verbs and nouns as demonstrated in Figure~\ref{fig:dataset}.
\vspace{-0.05in}
\subsection{Object-centric Representation} 
\vspace{-0.05in}
Given a video with $T$ frames, we first perform object detection on these video frames, using a detector which detects hands and generic candidate constituent objects. The object detector is trained on the set of all objects in the train split of the dataset as one class, and all hands in the training data as a second class. Assume that we have detected $N$ instances including the hands and the objects manipulated by the hands in the scene, we then perform multi-object tracking to find correspondences between boxes in different video frames. We extract two types of feature representation for each box: (a) bounding box coordinates; and (b) an object identity feature. Both of these features are designed for compositional generalization and avoiding object appearance bias. 

\vspace{-0.05in}
\myparagraph{Bounding box coordinates.} One way to represent an object and its movement is to use its location and shape. We use the center coordinate of each object along with its height and width as a quadruple, and forward it to a   Multi-Layer Perceptron (MLP), yielding a $d$-dimensional feature. Surprisingly, this simple representation alone turns out to be highly effective in action recognition.

\vspace{-0.05in}
\myparagraph{Object identity embedding.} In addition to the object coordinate feature, we also use a learnable $d$-dimensional embedding to represent the identities of objects and subjects. We define three types of embedding: (i) \textit{subject} (or \textit{agent}) embedding, \ie, representing hands in an action; (ii) \textit{object} embedding, \ie, representing the objects involved in the action; (iii) \textit{null} embedding, \ie, representing dummy boxes irrelevant to the action. The three embeddings are initialized from an independent multivariate normal distribution. The identity embedding can be concatenated together with box coordinate features as the input to our model. Since the identity (category) of the instances is predicted by the object detector, we can combine coordinate features with embedding features accordingly. We note that these embeddings do not depend on the appearance of input videos.

We find that combining the box coordinate feature with the identity feature significantly improves the performance of our model. Since we are using a \textit{general object} embedding for all kinds of objects, this helps the model to generalize across different combinations of verbs and nouns in a compositional action recognition setting. 

\textit{Robustness to Unstable Detection.} In cases where object detector is not reliable, where the number of detected objects is larger than a fix number $N$, we can perform object configuration search during inference. Each time we randomly sample $N$ object tracklets and forward them to our model. We perform classification based on the most confident configuration which has the highest score. However, in our current experiments, we can already achieve significant improvement without this process.

\vspace{-0.05in}
\subsection{Spatial-temporal interaction reasoning} 
\vspace{-0.05in}
Given $T$ video frames and $N$ objects per frame, we denote the set of object features as $X=(x_1^1, ..., x_N^1, x_1^2, ..., x_N^2, ...,x_N^T)$, where $x_i^t$ represents the feature of object $i$ in frame $t$. Our goal is to perform spatial-temporal reasoning in $X$ for action recognition. As illustrated in Figure~\ref{fig:method}(a), we first perform spatial interaction reasoning on objects in each frame, then we connect these features together with temporal interaction reasoning. 

\vspace{-0.05in}
\myparagraph{Spatial interaction module.} We perform spatial interaction reasoning among the $N$ objects in each frame. For each object $x_i^t$, we first aggregate the features from the other $N-1$ objects by averaging them, then we concatenate the aggregated feature with $x_i^t$. This process can be represented as,
\begin{align}
\vspace{-0.03in}
\label{eq:stm}
f (x_i^t) = {\rm ReLU}(W_{f}^{T}[x_i^t, \frac{1}{N-1} \sum_{j \neq i} x_j^t]), 
\vspace{-0.03in}
\end{align}
where $[,]$ denotes concatenation of two features in the channel dimension and $W_{f}^{T}$ is learnable weights implemented by a fully connected layer. We visualize this process in Figure~\ref{fig:method}(b) in the case of $N=3$.

\vspace{-0.05in}
\myparagraph{Temporal interaction module.} Given the aggregated feature of objects in each frame, we perform temporal reasoning on top of the features. As tracklets are formed and obtained previously, we can directly link objects of the same instance across time. Given objects in the same tracklet, we compute the feature of the tracklet as $g(x_i^1,...,x_i^T)$:  We first concatenate the object features, then forward the combined feature to another MLP network. Given a set of temporal interaction results, we aggregate them together for action recognition as, 
\begin{align}
\vspace{-0.03in}
p(X) = W_{p}^{T} h(\{g(x_i^1,...,x_i^T)\}^N_{i=1}), 
\vspace{-0.03in}
\end{align}
where $h$ is a function combining and aggregating the information of tracklets. In this study, we experiment with two different approaches to combine tracklets: (i) Design $h$ as a simple averaging function to prove the effectiveness of our spatial-temporal interaction reasoning. (ii) Utilize non-local block~\cite{xiaolongwang2017nonlocal} as the function $h$. The non-local block encodes the pairwise relationships between every two trajectory features before averaging them. In our implementation, we adopt three non-local blocks succeeded by convolutional kernels. 
We use $W_{p}$ as our final classifier with cross-entropy loss.
\vspace{-0.05in}

\myparagraph{Combining video appearance representation.} Besides explicitly modeling the transformation of relationships of subjects and objects, our spatial-temporal interaction model can be easily combined with any video-level appearance representation. The presence of appearance features helps especially the action classes without prominent inter-object dynamics. To achieve this, we first forward the video frames to a 3D ConvNet. We follow the network backbone applied in~\cite{wang2018videos}, which takes $T$ frames as input and extracts a spatial-temporal feature representation. We perform average pooling across space and time on this feature representation, yielding a $d$-dimensional feature. Video appearance representations are concatenated with object representations $h(\{g(x_i^1,...,x_i^T)\}^N_{i=1})$, before fed into the classifier.  
\vspace{-0.05in}
\section{The Something-Else Task}
\vspace{-0.05in}
\label{sec:dataset}
To present the idea of compositional action recognition, we adopt the Something-Something V2 dataset~\cite{goyal2017something} and create new annotations and splits within it. We name the action recognition on the new splits as the ``Something-Else task''.

The Something-Something V2 dataset contains 174 categories of common human-object interactions. Collected via Amazon Mechanical Turk in a crowd-sourced manner, the protocol allows turkers to pick an action category (\emph{verb}), perform and upload a video accordingly with arbitrary objects (\emph{noun}). The lack of constraints in choosing the objects naturally results in large variability in the dataset. There are $12,554$ different object descriptions in total. The original split does not consider the distribution of the objects in the training and the testing set, instead, it asserts that the videos recorded by the same person are in either training or testing set but both. While this setting reduces the environment and individual bias, it ignores the fact that the combination of verbs and nouns presented in the testing set may have been encountered in the training stage. The high performance obtained in this setting might indicate that models have learned the actions coupled by typical objects occurring, yet does not reflect the generalization capacity of models to actions with novel objects. 

\vspace{-0.05in}
\myparagraph{Compositional Action Recognition.} In contrast to randomly assigning videos into training or testing sets, we present a compositional action recognition task. In our setting, the combinations of a verb (action) and nouns in the training set do not exist in the testing set. We define a subset of \emph{frequent object categories} as those appearing in more than 100 videos in the dataset. We split the \emph{frequent object categories} into two disjoint groups, $\mathcal{A}$ and $\mathcal{B}$. Besides objects, action categories are divided into two groups ${1}$ and ${2}$ as well. In~\cite{goyal2017something} these categories are organized hierarchically, \eg, \emph{``moving something up''} and \emph{``moving something down''} belong to the same super-class. We randomly assign each action category into one of two groups, and at the same time enforce that the actions belonging to the same super-class are assigned into the same group. 

Given the splits of groups, we combine action group $1$ with object group $\mathcal{A}$, and action group ${2}$ with object group $\mathcal{B}$, to form the training set, termed as $1\mathcal{A} + 2\mathcal{B}$. The validation set is built by flipping the combination into $1\mathcal{B} + 2\mathcal{A}$. Different combinations of verbs and nouns are thus divided into training \emph{or} testing splits in this way. The statistics of the training and the validation sets under the compositional setting are shown in the second row of Table~\ref{tab:stats}.  

\begin{figure*}[!t]
\centering
\includegraphics[width=.98\linewidth]{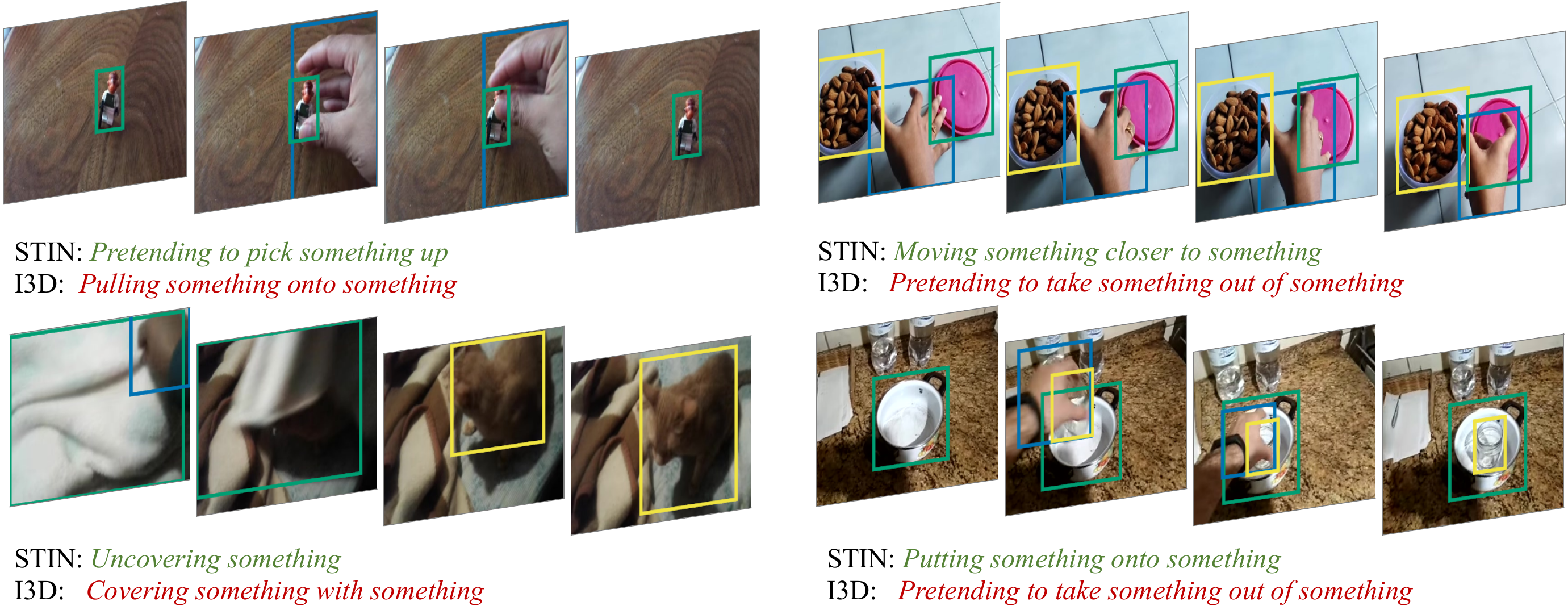}
\caption{Predictions of STIN and I3D models. Correct predictions are in green, incorrect in red. STIN can keep tracking the relations between subjects and objects as they change over time in complicated actions.}
\vspace{-0.1in}
\label{fig:visualization}
\end{figure*}

\vspace{-0.05in}
\begin{table}[t]
\centering
\small
\begin{tabular}{l|r|r r}
\multicolumn{1}{c|}{Task Split} & \multicolumn{1}{c|}{\# Classes} & \multicolumn{1}{c}{Training} & \multicolumn{1}{c}{Validation} \\ 
\shline
Original & 174 & 168,913 & 24,777 \\
\hline
Compositional & 174 & 54,919 & 57,876 \\
\hline
FS-Base & 88 & 112,397 & 12,467 \\
FS-Novel 5-S & 86 & 430 & 49,822 \\
FS-Novel 10-S & 86 & 860 & 43,954 \\
\end{tabular}
\caption{\textbf{Comparison and statistics of various tasks} on the Something-Something V2. FS: few-shot; $n$-S: $n$-shot.}
\label{tab:stats}
\vspace{-0.15in}
\end{table}

\myparagraph{Few-shot Compositional  Action Recognition.} The compositional split challenges the network to generalize over object appearance. We further consider a few-shot dataset split setting indicating how well a trained action recognition model can generalize to novel action categories with only a few training examples. We assign the action classes in the Something-Something V2 dataset into a \textit{base} split and a \textit{novel} split, yielding 88 classes in the base set and 86 classes in the novel set. We randomly allocate $10\%$ of the videos from the base set to form a validation set and the rest of the videos as the base training set. We then randomly select $k$ examples for each category in the novel set whose labels are present in the training stage, and the remaining videos from the novel set are designated as the validation set. We ensure that the object categories in $k$-shot training videos do not appear in the novel validation set. In this way, our few-shot setting additionally challenges models to generalize over object appearance. We term this task as few-shot compositional recognition. We set $k$ to $5$ or $10$ in our experiments. The statistics are shown in Table~\ref{tab:stats}. 
\vspace{-0.05in}
\myparagraph{Bounding-box annotations.} We annotated 180,049 videos of the Something-Something V2 dataset. For each video, we provide a bounding box of the hand (hands) and objects involved in the action. In total, 8,183,381 frames with 16,963,135 bounding boxes are annotated, with an average of 2.41 annotations per frame and 94.21 per video. Other large-scale video datasets use bounding box annotation, in applications involving human-object interaction~\cite{damen2018scaling}, action recognition~\cite{gu2018ava}, and tracking~\cite{real2017youtube}.
\vspace{-0.05in}

%% file: 4_experiments.tex
\section{Experiments}
\label{sec:exp}
\vspace{-0.1in}
We perform experiments on the two proposed tasks: compositional action recognition and few-shot compositional action recognition. 
\vspace{-0.05in}
\subsection{Implementation Details} 
\vspace{-0.1in}
\myparagraph{Detector.} We choose Faster R-CNN~\cite{ren2015faster,wu2019detectron2} with Feature Pyramid Network (FPN)~\cite{lin2017feature} and ResNet-101~\cite{he2016deep} backbone. The model is first pre-trained with the COCO~\cite{lin2014microsoft} dataset, then finetuned with our object box annotations on the Something-Something dataset. During finetuning, only two categories are registered for the detector: \emph{hand} and \emph{object} involved in action. The object detector is trained with the same split as the action recognition model.
We set the number of objects in our model as 4. If fewer objects are presented, we fill a zero vector to represent the object.\par
\vspace{-0.05in}
\myparagraph{Tracker.} Once we have the object detection results, we apply multi-object tracking to find correspondence between the objects in different frames. The multi-object tracker is implemented based on minimalism to keep the system as simple as possible. Specifically, we use the Kalman Filter~\cite{kalman1960filter} and Kuhn-Munkres (KM) algorithm~\cite{kuhn1955hungarian} for tracking objects as~\cite{bewley2016simple}. At each time step, the Kalman Filter predicts plausible whereabouts of instances in the current frame based on previous tracks, then the predictions are matched with single-frame detections by the KM algorithm. 
\vspace{-0.1in}
\subsection{Setup}
\vspace{-0.1in}
\myparagraph{Training details.}
The MLP in our model contains 2 layers. We set the dimension of MLP outputs $d=512$. We train all our models for 50 epochs with learning rate 0.01 using SGD with 0.0001 weight decay and 0.9 momentum, the learning rate is decayed by the factor of 10 at epochs 35 and 45.

\vspace{-0.05in}
\paragraph{Methods and baselines.} The experiments aim to explore the effectiveness of different components in our Spatial-Temporal Interaction Networks for compositional action recognition, we compare the following models: 

\begin{itemize}
\vspace{-0.08in}
\setlength\parskip{-0.1em}
\item \textbf{STIN}: Spatial-Temporal Interaction Network with bounding box coordinates as input. Average pooling is used as aggregation operator $h$. 
\item \textbf{STIN + OIE}: STIN model not only takes box coordinates but also Object Identity Embeddings (OIE). 
\item \textbf{STIN + OIE + NL}: Use non-local operators for aggregation operator $h$ in STIN + OIE. 
\item \textbf{I3D}: A 3D ConvNet model with ResNet-50 backbone as  in~\cite{wang2018videos}, with state-of-the-art performance. 
\item \textbf{STRG}: Space-Time Region Graph (STRG) model introduced in~\cite{wang2018videos} with only similarity graph.
\item \textbf{I3D + STIN + OIE + NL}: Combining the appearance feature from the I3D model and the feature from the STIN + OIE + NL model by joint learning.
\item \textbf{I3D, STIN + OIE + NL}: A simple ensemble model combining the separately trained I3D model and the trained STIN + OIE + NL model. 
\item \textbf{STRG, STIN + OIE + NL}: An ensemble model combining the STRG model and the STIN + OIE + NL model, both trained separately.
\end{itemize}

Our experiments with STIN use either ground-truth boxes or the boxes detected by the object detector. The presented score is from a single clip in each video, which is a center cropped in time.

\paragraph{Visualization.} Figure~\ref{fig:visualization} visualizes examples of how our STIN model and I3D model performs. Our STIN model can keep tracking how the hand moves to understand the action whereas I3D is confused when the activity resembles other action class.

\vspace{-0.03in}
\subsection{Original Something-Something Split}
\vspace{-0.05in}
We first perform our experiments on the original Something-something V2 split. We test our I3D baseline model and the STIN model with ground-truth object bounding boxes for action recognition. As shown in Table~\ref{tab:original}, our I3D baseline is much better than the recently proposed TRN~\cite{zhou2018temporal} model.
The result of our STIN + OIE model with ground-truth annotations is reported in Table~\ref{tab:original}. We can see that with only coordinates inputs, our performance is comparable with TRN~\cite{zhou2018temporal}. After combining with the I3D baseline model, we can improve the baseline model by $5\%$. This indicates the potential of our model and bounding box annotations even for the standard action recognition task.

\subsection{Compositional Action Recognition}
\vspace{-0.05in}
We further evaluate our model on the compositional action recognition task. 
We first experiment with using the ground-truth object bounding boxes for the STIN model, as reported in Table~\ref{tab:comp:gt}. To illustrate the difficulty of our compositional task, we also report the results on a ``shuffled'' split of the videos: We use the same candidate videos but shuffle them randomly and form a new training and validation set. Note that the number of training videos is the same as the compositional split. The performance of the I3D baseline sharply drops from the shuffled setting to compositional setting by almost $15\%$ in terms of top-1 accuracy. On the shuffled split, although our STIN model trails I3D, it performs better than I3D in the compositional split. By applying the Object Identity Embedding (OIE), we can improve the STIN model by $4.3\%$. This attests to the importance of explicit reasoning about the interactions between the agent and the objects. We can further combine our model with the I3D baseline: the joining of two models yields $7.8\%$ improvement over the baseline and the ensemble model significantly improves over the  appearance only model (I3D) by $11.3\%$.\par
\vspace{-0.03in}
Following, we build our model on object bounding boxes obtained via object detection and tracking and show its results in Table~\ref{tab:comp:det}. We observe that OIE still boosts the STIN model by $3.1\%$. By combining I3D with our model, we observe $1.4\%$ improvement over I3D%
\vspace{-0.03in}
We also see that by replacing the base network with STRG we obtain some improvement in performance over I3D. After combining the STRG model with our model (STRG, STIN + OIE + NL), we can still achieve a large relative improvement ($3.9\%$ better than STRG). This shows that our method is complementary to the existing graph model.

\begin{table}[t]
\centering
\small
\tablestyle{4pt}{1.05}
\begin{tabular}{l|x{22}x{22}}
\multicolumn{1}{c|}{model} & top-1 &  top-5 \\ [.1em]
\shline
TRN~\cite{zhou2018temporal} & 48.8 & 77.6  \\
TRN Dual Attention~\cite{xiao2019reasoning}  & 51.6 &  80.3  \\
TSM~\cite{lin2018temporal} & 61.7 & 87.4 \\
\hline
STIN + OIE & 48.4 & 78.7\\
I3D & 55.5 & 81.4 \\
I3D + STIN + OIE & 60.2 & 84.4 \\
\end{tabular}
\vspace{0.3em}
\caption{Results on the \textbf{original Something-something V2} dataset. Ground-truth annotations are applied with STIN.}
\vspace{-0.1in}
\label{tab:original}
\end{table}

\begin{table}[t]\centering
\subfloat[Compositional action recognition with \textbf{ground-truths}. \label{tab:comp:gt}]{
\tablestyle{2pt}{1.05}
\begin{tabular}{l|l|x{28}x{28}}
\multicolumn{1}{c|}{model}  & \multicolumn{1}{c|}{split}  & top-1 & top-5 \\
\shline
STIN & Shuffled & 54.0 & 79.6  \\
STIN & Compositional & 47.1 & 75.2 \\
STIN + OIE & Compositional & 51.3 & 79.3  \\
STIN + OIE + NL & Compositional & 51.4 & 79.3  \\
\hline
I3D & Shuffled & 61.7 & 83.5 \\
I3D & Compositional & 46.8 & 72.2  \\
I3D + STIN + OIE + NL  & Compositional & 54.6 &79.4  \\
I3D, STIN + OIE + NL  & Compositional &  58.1 & 83.2  \\
\end{tabular}}\hspace{3mm}
\subfloat[Compositional action recognition with \textbf{detections}. \label{tab:comp:det}]{
\tablestyle{2pt}{1.05}
\begin{tabular}{l|l|x{28}x{28}}
\multicolumn{1}{c|}{model} & \multicolumn{1}{c|}{split} & top-1 & top-5 \\
\shline
STIN & Compositional &34.1& 58.8 \\
STIN + OIE & Compositional &36.7& 62.2\\
STIN + OIE + NL & Compositional &37.2 & 62.4   \\
\hline
I3D & Compositional & 46.8 & 72.2 \\
STRG & Compositional & 52.3 & 78.3 \\
I3D + STIN + OIE + NL & Compositional & 48.2 & 72.6  \\
I3D, STIN + OIE + NL & Compositional & 51.5 & 77.1  \\
STRG, STIN + OIE + NL & Compositional & 56.2 & 81.3  \\
\end{tabular}} 
\caption{\textbf{Compositional action recognition} over 174 categories. }
\vspace{-0.15in}
\label{tab:comp}
\end{table}
\vspace{-0.03in}

\vspace{-0.03in}
\subsection{Few-shot Compositional Action Recognition}
\vspace{-0.05in}
For the few-shot compositional action recognition task, we have 88 \textit{base} categories and 86 \textit{novel} categories as described in Section~\ref{sec:dataset}. We first train our model with the videos from the base categories, then finetune on few-shot samples from the novel categories. We evaluate the model on the novel categories with more than 50k videos to benchmark the generalizability of our model. 
For finetuning, instead of following the $n$-way, $k$-shot setting in few-shot learning~\cite{finn2017model}, we directly fine-tune our model with all the novel categories. For example, if we perform 5-shot training, then the number of training examples is $86{\times}5{=}430$. During the fine-tuning stage, we randomly initialize the last classification layer and train this layer while fixing all other layers. We train the network for 50 epochs with a fixed learning rate of $0.01$. We perform both 5-shot and 10-shot learning in our experiment.

We report our results with ground-truth object boxes in Table~\ref{tab:few:gt}. 
We can see that our full model (STIN+OIE+NL) outperforms the I3D model by almost $6\%$ in both 5-shot and 10-shot learning setting, even though our approach trails I3D on the validation set in \textit{base} categories. This indicates that the I3D representation can easily overfit to object appearance while our model generalizes much better.  We also observe the OIE and non-local block individually and cooperatively boost the few-shot performance. When combining with I3D with the model ensemble, we achieve $12.2\%$ improvement on 5-shot and $13.9\%$ on the 10-shot setting. 
The results with object detection boxes are shown in Table~\ref{tab:few:det}. Although the best model STIN+OIE+NL trails I3D on \textit{base} evaluation by a notably large margin, the performance in the few-shot setting is much closer.
When combining our model with the I3D model, joint learning yields $1.9\%$ improvement and model ensemble yields $5.5\%$ improvement in the 5-shot setting. We observe similar improvement in the 10-shot setting ($5.9\%$). By replacing the I3D base network with STRG, our method (STRG, STIN + OIE + NL) still gives large improvement over STRG ($4.3\%$ in 5-shot setting and $4.7\%$ in 10-shot).

\begin{figure}[t]
\centering
\includegraphics[width=0.35\textwidth]{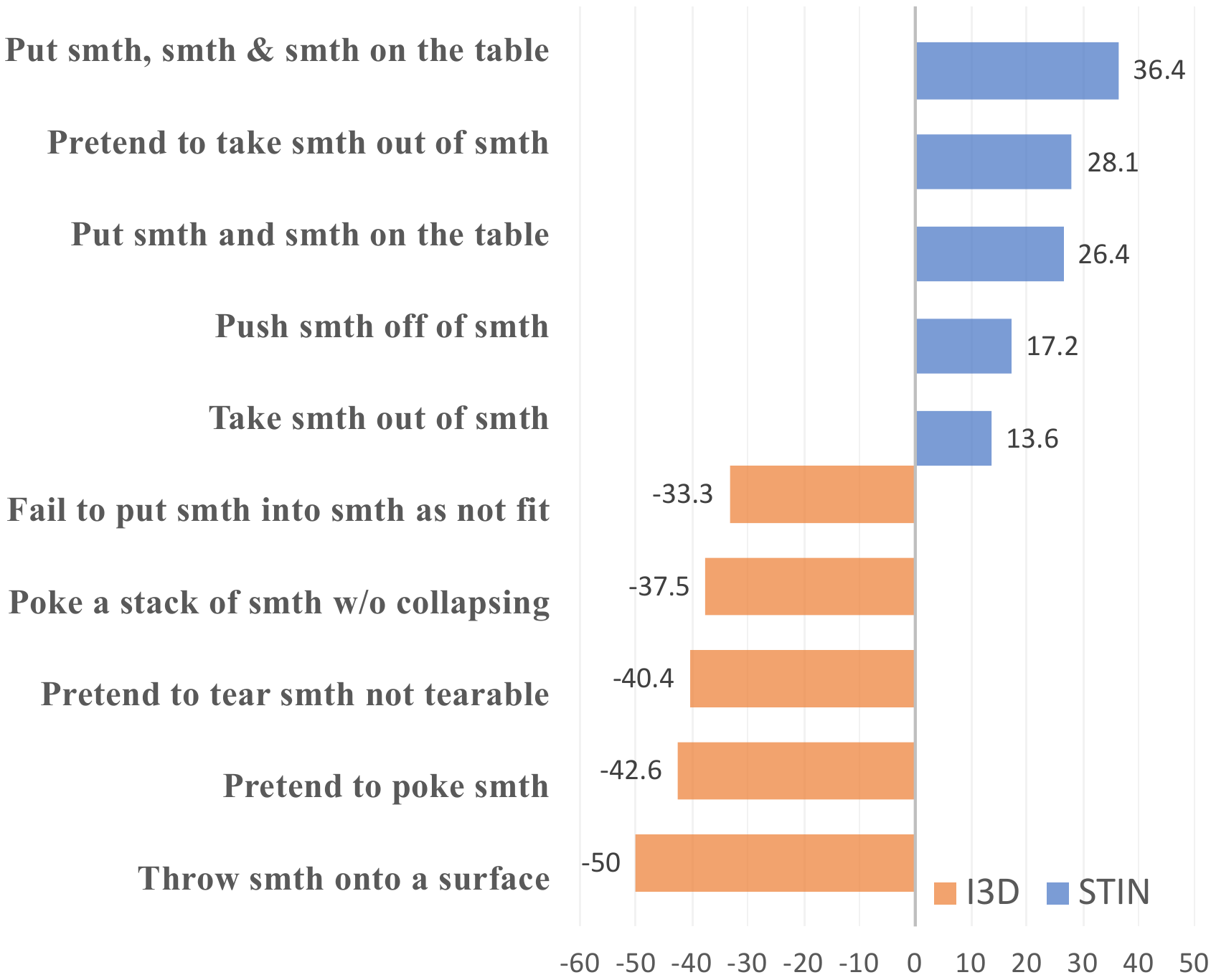}%
 \caption{Top categories on which STIN surpasses or trails I3D, the numbers represent the difference in accuracy of the both models.}
 \vspace{-0.13in}
\label{fig:category}

\end{figure}
\vspace{-0.05in}
\subsection{Ablations}
 \vspace{-0.05in}
\paragraph{One-object training.}
We push the compositional setting to an extreme, where we only select the videos where actions are interacting with the object category ``box'' for training ($6,560$ videos in $166$ action categories). The rest of the videos are the validation set (170K videos). The objective of this experiment is to examine the generalizability of our STIN model, even when the training set is strongly biased toward one type of object. 

The results are summarized in Table~\ref{tab:comp:oneobj}. Our model with ground-truth boxes almost doubles the I3D performance. Our model with detection boxes is also $5.6\%$ better than I3D. This attests to the advantage of our model in terms of generalizability across different object appearances. 
\begin{table}[t]\centering
\subfloat[Few-shot compositional setting with \textbf{ground-truths}. \label{tab:few:gt}]{
\tablestyle{2pt}{1.05}
\begin{tabular}{l|x{28}x{28}|x{28}x{28}}
\multicolumn{1}{c|}{~}  &  \multicolumn{2}{c|}{\emph{base}}  &  \multicolumn{2}{c}{\emph{few-shot}} \\
\multicolumn{1}{c|}{model}  & top-1 & top-5  & 5-shot & 10-shot \\
\shline
STIN & 65.7 & 89.1 & 24.5 & 30.3\\
STIN + OIE &69.5 &91.4 & 25.8 & 32.9 \\
STIN + OIE + NL & 70.2 & 91.4 & 27.7& 33.5 \\
\hline
I3D & 73.6 & 92.2 & 21.8 & 26.7 \\
I3D + STIN + OIE + NL  & 80.6 & 95.2 & 28.1 & 33.6 \\
I3D, STIN + OIE + NL  & 81.1& 96.0 & 34.0& 40.6\\
\end{tabular}}\hspace{3mm}
\subfloat[Few-shot compositional setting with  \textbf{detections}.\label{tab:few:det}]{
\tablestyle{2pt}{1.05}
\begin{tabular}{l|x{28}x{28}|x{28}x{28}}
\multicolumn{1}{c|}{~}  &  \multicolumn{2}{c|}{\emph{base}}  &  \multicolumn{2}{c}{\emph{few-shot}} \\
\multicolumn{1}{c|}{model}  & top-1 & top-5  & 5-shot & 10-shot \\
\shline
STIN & 54.0& 78.9& 14.2 & 19.0 \\
STIN + OIE & 58.2 & 82.6& 16.3& 20.8 \\
STIN + OIE + NL & 58.2 & 82.6 & 17.7 & 20.7 \\
\hline
I3D & 73.6 & 92.2 & 21.8 & 26.7 \\
STRG &  75.4 & 92.7  &  24.8 & 29.9   \\
I3D + STIN + OIE + NL  & 76.8 & 93.3 &23.7 & 27.0 \\
I3D, STIN + OIE + NL  & 76.1&  92.7   & 27.3 & 32.6 \\
STRG, STIN + OIE + NL  &78.1 & 94.5  & 29.1 & 34.6\\
\end{tabular}}
\vspace{0.05in}
\caption{\textbf{Few-shot compositional action recognition} on \textit{base} categories and \textit{few-shot novel} categories. We show results with (a) ground-truth bounding boxes and (b) object detection boxes. }
\vspace{-0.05in}
\label{tab:ablations}
\end{table}

\begin{table}[t]\centering
\tablestyle{2pt}{1.05}
\begin{tabular}{l|x{55}|x{28}x{28}}
\multicolumn{1}{c|}{model}  & split  & top-1 & top-5 \\
\shline
STIN + OIE (GT) & Compositional & 28.5& 54.1  \\
STIN + OIE (Detector) & Compositional  & 20.3 & 40.4 \\
\hline
I3D & Compositional & 14.7 & 34.0  \\
\end{tabular}
\vspace{0.05in}
\caption{\textbf{One-class} compositional action recognition. The model is trained on videos with only one object class: ``box''.}
\vspace{-0.10in}
\label{tab:comp:oneobj}
\end{table}
\vspace{-0.05in}
\paragraph{Category analysis.} We compare the performance difference between our STIN and the I3D model for individual action categories. We visualize the five action categories that STIN surpasses or trails by the largest margin compared to the I3D model in Figure~\ref{fig:category}.
\emph{A priori}, actions that are closely associated with the transformation of the object's geometric relations should be better represented by the STIN model than I3D. We can see that the actions in which STIN outperforms I3D by the largest margin are the ones that directly describe the movements of objects, such as \emph{``put something''} and \emph{``take something''}. On the other hand, STIN fails when actions are associated more with the changes in terms of the intrinsic property of an object, such as \emph{``poking''} and \emph{``tearing''}.

%% file: 5_conclusion.tex
\vspace{-0.05in}
\section{Conclusion}
\vspace{-0.05in}
\label{sec:conclude}
Motivated by the appearance bias in current activity recognition models, we propose a new model for action recognition based on sparse semantically grounded subject-object graph representations. We validate our approach on novel compositional and few shot settings in the Something-Else dataset; our model is trained with new constituent object grounding annotations
Our STIN approach models the interaction dynamics of objects composed in an action and outperforms all baselines. 